\documentclass[runningheads,a4paper]{llncs}

\usepackage{amssymb}
\usepackage{graphicx}
\usepackage[namelimits]{amsmath}
\usepackage[linesnumbered,ruled,vlined]{algorithm2e}

\usepackage{url}
\urldef{\mailsa}\path|{hwang, qifu, howang, syue}@lincoln.ac.uk|
\urldef{\mailsb}\path|jgpeng@gzhu.edu.cn|
\urldef{\mailsc}\path|syue@lincoln.ac.uk|
\newcommand{\keywords}[1]{\par\addvspace\baselineskip
\noindent\keywordname\enspace\ignorespaces#1}

\usepackage[utf8]{inputenc}

\usepackage{cite}

\usepackage{nameref,hyperref}

\usepackage{lastpage,fancyhdr,graphicx}

\usepackage{color}

\definecolor{Gray}{gray}{.25}

\usepackage{graphicx}

\begin{document}
\mainmatter

\title{Constant Angular Velocity Regulation for Visually Guided Terrain Following }

\titlerunning{Constant Angular Velocity Terrain Following}
\author{Huatian Wang\textsuperscript{1},
Qinbing Fu\textsuperscript{1,2},
Hongxin Wang\textsuperscript{1},\\
Jigen Peng\textsuperscript{2,3},
and Shigang Yue\textsuperscript{1,3}}
\authorrunning{Constant Angular Velocity Terrain Following}


\institute{1 The Computational Intelligence Lab(CIL), School of Computer Science, \\ University of Lincoln, Lincoln, LN6 7TS, UK\\
\mailsa\\
2 School of Mathematics and Information Science, Guangzhou University, \\
Guangzhou, 510006, China\\
\mailsb\\
3 Machine Life and Intelligence Research Center, Guangzhou University, \\
Guangzhou,  510006, China\\
\mailsc\\
}

\maketitle

\begin{abstract}
Insects use visual cues to control their flight behaviours. By estimating the angular velocity of the visual stimuli and regulating it to a constant value, honeybees can perform a terrain following task which keeps the certain height above the undulated ground. For mimicking this behaviour in a bio-plausible computation structure, this paper presents a new angular velocity decoding model based on the honeybee's behavioural experiments. The model consists of three parts, the texture estimation layer for spatial information extraction, the motion detection layer for temporal information extraction and the decoding layer combining information from pervious layers to estimate the angular velocity. Compared to previous methods on this field, the proposed model produces responses largely independent of the spatial frequency and contrast in grating experiments. The angular velocity based control scheme is proposed to implement the model into a bee simulated by the game engine Unity. The perfect terrain following above patterned ground and successfully flying over irregular textured terrain show its potential for micro unmanned aerial vehicles' terrain following.

\end{abstract}

\section{Introduction}
The detection of visual motion has been researched for decades to understand how insects, like honeybees and locusts, use visual information to guide their flight behaviours\cite{Sri1996, Seidl1982, wang2018directionally}. Executing a visually guided terrain following is one of the most challenging parts in visual flight control for flying insects. To accomplish this, the insect might either measure the flight speed and adjust the height accordingly, or estimate the flight height directly. However, it is hard for their tiny brains, with insufficient computation resources, to measure the flight speed and the distance to surface. So a much simpler strategy, holding the angular velocity (the ratio of forward speed and distance) constant, is taken by honeybees for this visual control task \cite{Sri1997, franceschini2007bio, serres2017optic}.

The angular velocity can be measured by the angular displacement $\Delta \phi$ in a small time interval $\Delta t$, that is $\omega = \frac{\Delta \phi}{\Delta t}$. In terrain following scenario, denoting $v_x$ as the forward flight speed and $d$ as the distance to the surface, the angular velocity of the image motion perceived by ventral part of the compound eyes can also be expressed as $\omega = \frac{v_x}{d}$. If the forward speed is maintained by a constant forward thrust, then the flight altitude will change automatically as the distance to ground varies by regulating the angular velocity to a constant value. Problems therefore arise as how insect estimates the angular velocity and keeps it constant.

Biological experiments show that honeybees can fly in the central path of the patterned tunnel by balancing the angular velocity of the image motion on both eyes. What's more, the estimation of the angular velocity is largely independent of the spatial frequency and the contrast \cite{Sri1997, Baird2005}. Further, the spike recordings also show that the responses of some descending neurons in the honeybee’s central nerve cord grow as the angular velocity of the image motion increases \cite{Ibb2001, Ibb2017}. Both indicate that honeybees are capable of estimating the angular velocity. However, the neural mechanism underlying this ability is still elusive.

\begin{figure}[htb]
\centering
\includegraphics [width=60mm]{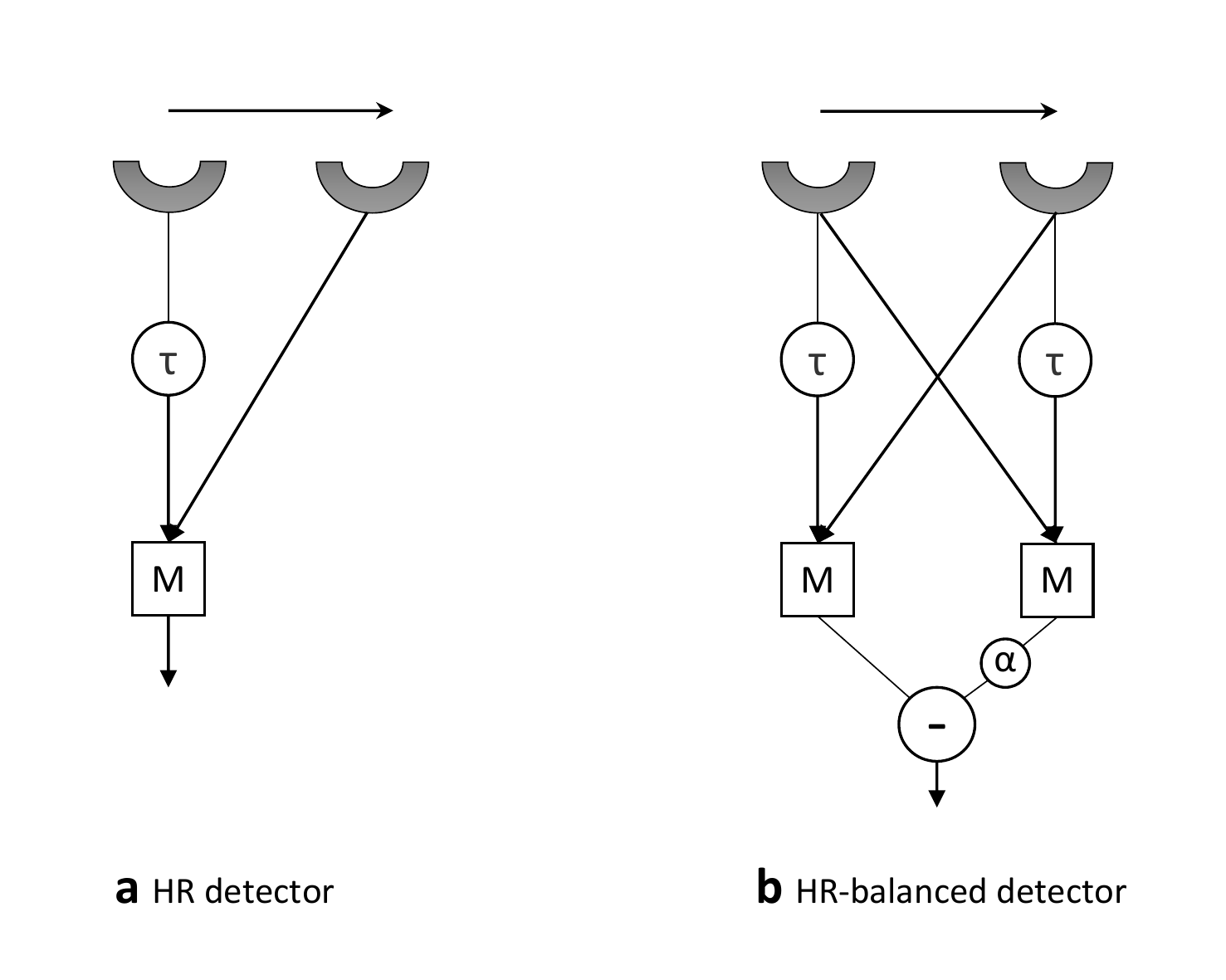}
\caption{HR detector and HR-balanced detector. (a) HR detector uses the multiplication (M) of the delayed signal from left photoreceptor and the non-delay signal from right to enhance the response of a preferred direction motion \cite{HR1956}. (b) HR-balanced detector can detect both preferred and opposite direction motions \cite{Zanker1999}.}
\label{fig1}
\end{figure}

Due to the limitation of the computation ability of insect's tiny brain, the differential techniques, matching or feature-based approaches \cite{fleet2012measurement} which are widely used in computer vision are not working here. Hassenstein and Richardt propose a classic elementary motion detecting model describing the mechanism of the motion sensing for animals \cite{HR1956}. The Hassenstein-Reichardt (HR) motion detector produces a much higher response when a progressive motion movement presents (see Fig. \ref{fig1}(a)). HR-balanced detector which has a mirror symmetrical structure with a balance parameter can detect motion from both directions\cite{Zanker1999} (see Fig. \ref{fig1}(b)).

An angular velocity sensor based on the HR model is designed \cite{ruffier2005optic} to accomplish visual guided aircraft terrain following by Franceschini and Ruffier \cite{ruffier2015optic, franceschini2007bio}. However, both the HR model and the HR-balanced model are tuned for particular temporal frequency (number of gratings passed over the photoreceptor per second) rather than angular velocity\cite{Zanker1999}. So the output of their sensor shows a large variance when tested by patterned ground in flight \cite{ruffier2005optic}. Therefore, an angular velocity specific model is needed to improve the terrain following performance.

According to numerical calculations, Zanker et al.\cite{Zanker1999} point that the ratio of two HR-balanced detectors is angular velocity tuned. Based on this idea, Cope et al. \cite{Cope2016} propose C-HR model, using the ratio of two HR-balanced detectors with different temporal delays. Their model perform well especially around 100 $^\circ/s$. However, this is different from the fact that the honeybees usually keep a constant angular velocity around 300 $^\circ/s$ \cite{Baird2005}. Riabinina and Philippides\cite{Ria2009} build up the R-HR model, using a fully temporal dependent channel as the denominator to produce angular velocity tuned responses.  But the response of their model slightly depend on the spatial frequency of the moving grating and the dependence increases as the angular velocity grows. Wang et al.\cite{Wang2018} propose a model based on the neural structure of Drosophila's visual system which compares the results from detectors with different sampling rates, to get a spatial independent response. Nevertheless, the independence still needs to be more significant before guiding the terrain following.

Previous mentioned models use ratio of two channels to get a angular velocity tuned response, but the spatial independence of the response is not strong enough to reproduce honeybees' flight behaviours. At the same time, the performance is affected when the angular velocity is small or large due to the division. To address this issue, this paper presents a new model using texture estimation layer to estimate the spatial frequency of the image. The proposed model preforms better by combining both texture and temporal information to decode the angular velocity of the image motion.

\section{Methods}

\subsection{Angular Velocity Decoding Model}
\begin{figure}[htb]
\centering
\includegraphics [width=110mm]{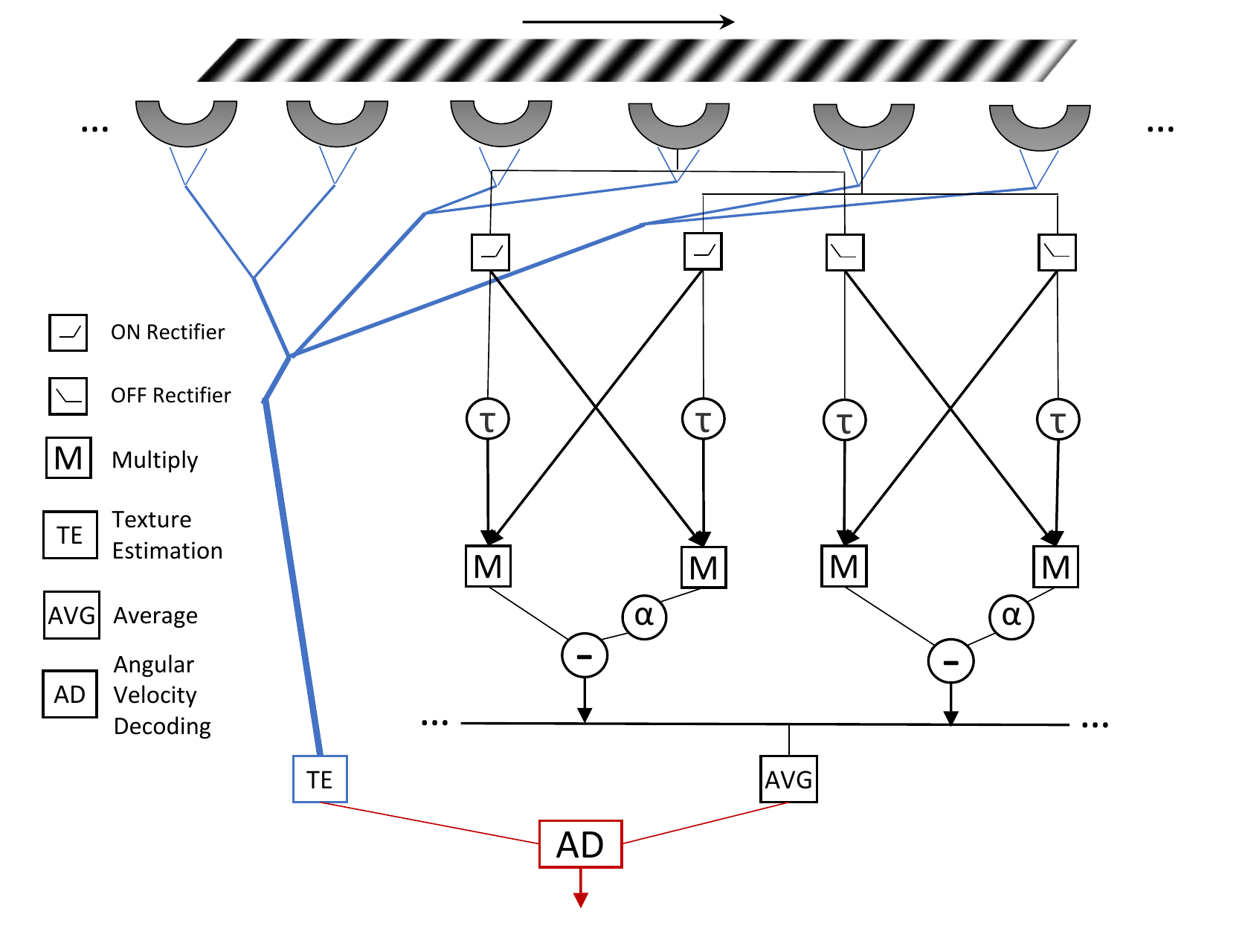}
\caption{The proposed Angular Velocity Decoding Model. The visual information of the grating movement is received by photoreceptors. The global spatial frequency and image contrast information is estimated by texture estimation part, and the temporal information is processed by motion detectors. Angular velocity decoding layer combines both texture and temporal information to estimate the angular velocity. }
\label{fig2}
\end{figure}

The model mainly contains 3 parts, the texture estimation part for spatial information extraction, the motion detection part for temporal information extraction and the decoding part for angular velocity estimation. The structure of the model is shown in Fig. \ref{fig2} explaining how these parts are connected with each other. In the proposed model, every motion detector receives the light intensity change from the neighbouring photoreceptors. And the light intensity change is separated into ON and OFF pathways and then processed by two HR-balanced detectors. The texture information and the motion information from the average of the detectors across the whole vision field are combined. Then angular velocity is decoded from this composite information.

\subsubsection{(1) Texture estimation layer.}

The simulated input signals received by ventral part of the compound eye are processed by a texture estimation layer where the image contrast and the spatial frequency of the gratings are estimated by the light intensities of different locations. This is based on a hypothesis that insects can have a basic sense of complexity of the texture. And the estimation method requires only low computation ability to give the global texture information of the spatial frequency and image contrast. Here we use Michelson contrast which is defined as the following:

\begin{equation}
C = \frac{I_{max}-I_{min}}{I_{max} + I_{min}}
\label{eq2}
\end{equation}
where the $I_{max}$ and $I_{min}$ ($I_{max}, I_{min} \geq 0$) indicate the highest and the lowest light intensities of the input signal in vision filed.

Then in order to decrease the cost of computing, binarization of the input image is performed using the relative intensity threshold $I_{thre} = (I_{max} - I_{min})/2$. The spatial frequency is estimated by counting the number of boundary lines of the binary image in whole visual field. This simple method works well for sine-wave and check-board gratings in our simulations. For more complex background, boundary number also indicates the complexity to some extent.

\subsubsection{(2) Motion detection layer.}

Motion detection layer mainly captures the motion information. The input image frames are first processed by the lamina layer where the light intensity change, which insects interest more than the intensity itself, are computed to get the primary information of visual motion. The output of a cell in this layer is defined by the following:
\begin{equation}
P(x,y,t) = I(x,y,t)-I(x,y,t-1) + \sum_{i =1}^{m}p_iP(x,y,t-i)
\label{eq3}
\end{equation}
where $P(x,y,t)$ corresponds to the luminance change of pixel (x,y) at time t; $m$ denotes the maximum number of the time steps that the persistence of the luminance change can last and the persistence coefficients $p_i  \in (0,1)$ is defined as following equation respectively \cite{yue2006collision}:
\begin{equation}
p_i= (1+e^{\mu i})^{-1}
\label{eq4}
\end{equation}

Then the luminance changes are separated into two pathways, ON pathway and OFF pathway\cite{yue2017modeling,fu2018shaping}. Specifically, the ON pathway deals with light intensity increments; whilst the OFF pathway processes brightness decrements. Denoting $f^+ = \max(0, f)$ and $f^- =\min(0,f)$, then we can express the outputs of the cells in this two pathways as following:
\begin{equation}
\begin{split}
P^{ON}(x,y,t) = P^+(x,y,t),  P^{OFF}(x,y,t) = P^-(x,y,t).
\end{split}
\label{eq5}
\end{equation}

Considering the delay of the visual signals received by neighbouring cells, we denote $D^{ON}(x,y,t)$ and $D^{OFF}(x,y,t)$ as the output of the ON and OFF detectors for horizontal motion which are computed by following the structure of the HR-balanced detector (Fig. \ref{fig1}(b)). Using a pure time delay of magnitude $\tau$, then we have the following expression:
\begin{equation}
\begin{split}
D^{ON}(x,y,t) = & P^+(x,y,t-\tau)\cdot P^+(x,y+1,t) \\
 & -\alpha P^+(x,y,t)\cdot P^+(x,y+1,t-\tau)
\end{split}
\label{eq6}
\end{equation}
where $\alpha$ is chosen from Zanker's paper\cite{Zanker1999} setting as 0.25  forming a partial balanced model. And $D^{OFF}(x,y,t)$ can be expressed similarly. Then the output of all motion detectors are averaged to give a response containing the motion information.

\subsubsection{(3) Angular velocity decoding layer.}
In order to decode angular velocity of the image motion from texture information and response of the motion detection layer, here we take only one detector for example to analyse how the response is affected by the input signals. Let $S_1, S_2$ denote the input signal of photoreceptor A (left) and B (right), and $S_1^D, S_2^D$ denote the temporal delayed signal of A and B, then according to the structure of HR-balanced detector (Fig. \ref{fig1}(b)), the response of the detector $R_0$ can be expressed as $\overline{S_1^D \cdot S_2-\alpha S_2^D \cdot S_1}$, where the bar means the response is averaged over a time period to remove fluctuation caused by oscillatory input.

If the input signals are simulated using the following sinusoidal grating frames \eqref{eq7} with spatial period $\lambda$ moving at an angular velocity $\omega$:
\begin{equation}
I(x,y,t) = (\sin(\frac{2\pi \omega}{\lambda} (t - \frac{\varphi  (y-1)}{\omega}))+ 1/C)/(1/C+1)
\label{eq7}
\end{equation}
where (x,y) denotes the location of the ommatidium, t indicates the time and $C \in (0,1]$ denotes the image contrast. Then we can get the output of each detector by \eqref{eq6}. It can be roughly expressed in theoretical \cite{Zanker1999} as the following equation:
\begin{equation}
\begin{split}
R_0 \approx \frac{1-\alpha}{(1+C)^2} + \frac{C^2}{2(1+C)^2}[\sin(\frac{2\pi  (\varphi-\tau \omega )}{\lambda}) - \alpha \sin(\frac{2\pi (\varphi+ \tau \omega )}{\lambda})].
\end{split}
\label{eq8}
\end{equation}

In fact, the angular velocity of the background moving is caused by the flying of the insects. Considering when a simulated bee performing terrain following, the consistency of the image motion speed in vision field helps us simplify the problem so that we can average the output signals from all ON and OFF detectors in visual field to get the final response $R(\omega,\lambda)$ which encodes the angular velocity.

However, it is hard to derive angular velocity directly from \eqref{eq8}. But we can decode the angular velocity information from the response $R(\omega,\lambda)$ using an approximation method. Though there is an inevitable fitting error, we can decrease it into an acceptable level if the fitting function is chosen well. One decoding function can be chosen as following to approximate the actual angular velocity:
\begin{equation}
\widehat{\omega} = a^*\widehat{\lambda}^{b^*}\frac{1+\widehat C}{2\widehat C} \sqrt{R}
\label{eq9}
\end{equation}
where $\widehat{\omega}$ denotes the decoded angular velocity, $\widehat{\lambda}$ is the estimated spatial period and $\widehat{C}$ is the estimated contrast from texture estimation layer. Parameters $a^*$ and $b^*$ can be learned by minimizing the difference from the ground truth using alternate iteration method:
\begin{equation}
(a^*,b^*) = \arg \min_{a,b}\ (\omega - a\lambda^b \frac{1+C}{2C}\sqrt{R(\omega,\lambda)}).
\label{eq10} 
\end{equation}

\subsection{Control Scheme for Automatic Terrain Following }
The proposed model can be used to simulate the automatic terrain following of the honeybees by maintaining a constant angular velocity. Using AVDM, we can estimate the angular velocity in flight. By regulating it accordingly to a constant value, the altitude will change automatically even without the exact altitude or forward speed in flight.

\begin{figure}
\centering
\includegraphics [width=120mm]{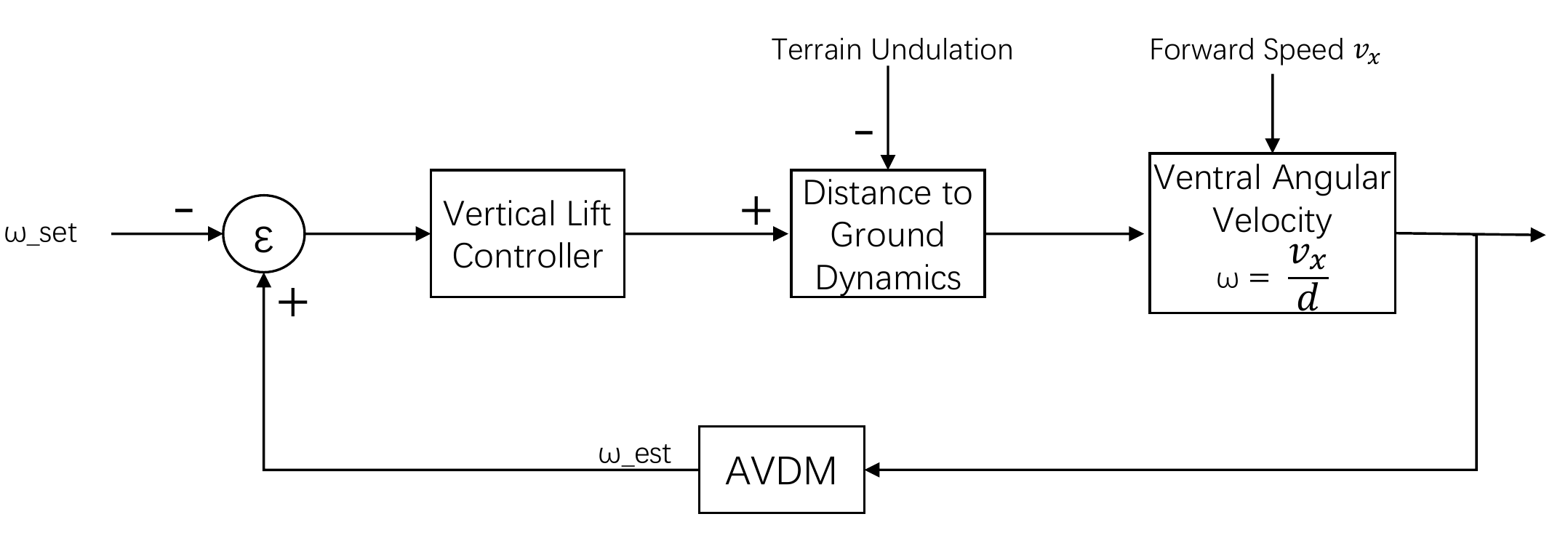}
\caption{ The AVDM-based close loop control terrain following scheme. The vertical lift controller is triggered by the difference $\varepsilon$ between preset angular velocity $\omega_{set}$ and the estimated angular velocity $\omega_{est}$. }
\label{fig3}
\end{figure}

The close loop control scheme for terrain following is given in Fig \ref{fig3}. For simplicity, we assume the forward flight speed is maintained  the same by a proper constant forward thrust. So we only need to adjust the vertical lift according to the difference between the preset angular velocity and the estimated value by AVDM to control the altitude correspondingly. Here, the preset angular velocity is also estimated by the AVDM in the beginning phase when the vertical lift is set to the same value as the gravity and where the ground is flat. After that, when the ventral angular velocity varies caused by terrain undulates, the vertical lift controller will change the lift according to the difference $\varepsilon$ between ventral angular velocity estimated and the preset value. If the difference $\varepsilon$ is positive, the lift will increase and vice versa.

During the terrain following approach, the vertical speed $v_z$ is relatively small, and the air resistance can be approximated as $f = k v_z$. Then the vertical dynamics can be described using the following differential equations:
\begin{equation}
m \frac{dv_z}{dt} = T - k v_z - mg
\end{equation}
\begin{equation}
T = \rho (\omega_{est} - \omega_{set})
\end{equation}
\begin{equation}
v_z = \frac{dz}{dt}
\end{equation}
where m is the mass of the simulated bee, g is the gravity acceleration and T is the vertical lift. Given the initial conditions, then the flight trajectory can be computed step by step. In our simulation, this process can be achieved by the physics engine of Unity.

\subsection{Parameter Setting}
Parameters of the proposed model and the control scheme are shown in TABLE \ref{table1}. Parameters are mainly tuned manually based on our empirical knowledge and stay the same in the following simulations unless particularly stated.
\begin{table}[htb]
\caption{Parameters of the model and the control scheme}
\centering
\begin{tabular}{c c c c}
\hline
Eq.  &   &   &  Parameters      \\ \hline
(2)  &   &   &  $ m = 10 $  \\
(3)  &   &   &  $ \mu = 1 $  \\
(5)  &   &   &  $\tau = 0.08 s$,  $\alpha = 0.25$       \\
(6)  &   &   &  $\varphi = 2^\circ$    \\
(8)  &   &   &  $a^* = 48.84$, $b^* =1$ \\
(10)  &   &   &  $k = 0.1$, $g =9.81$ \\
(11)  &   &   &  $\rho = 0.04$ \\ \hline
\end{tabular}
\label{table1}
\end{table}

\section{Experiments and Results}
Within this section, we present the experiments and results. The proposed model is first tested by synthetic grating stimuli to show its spatial independence in Matlab ($\copyright$ The MathWorks, Inc.). Then the model is implemented into a virtual bee using Unity ($\copyright$ Unity Technologies) to simulate the automatic terrain following behaviours of honeybees.

\subsection{Moving grating experiments}
In the first kind of experiments, we aimed to inspect the spatial frequency independence of the proposed angular velocity decoding model. Therefore, the spatial period of the grating is chosen widely from 12$^\circ$ to 72$^\circ$  and the model is tested by these gratings moving at different angular velocities to get a general result.

\begin{figure}[htb]
\centering
\includegraphics [width=60mm]{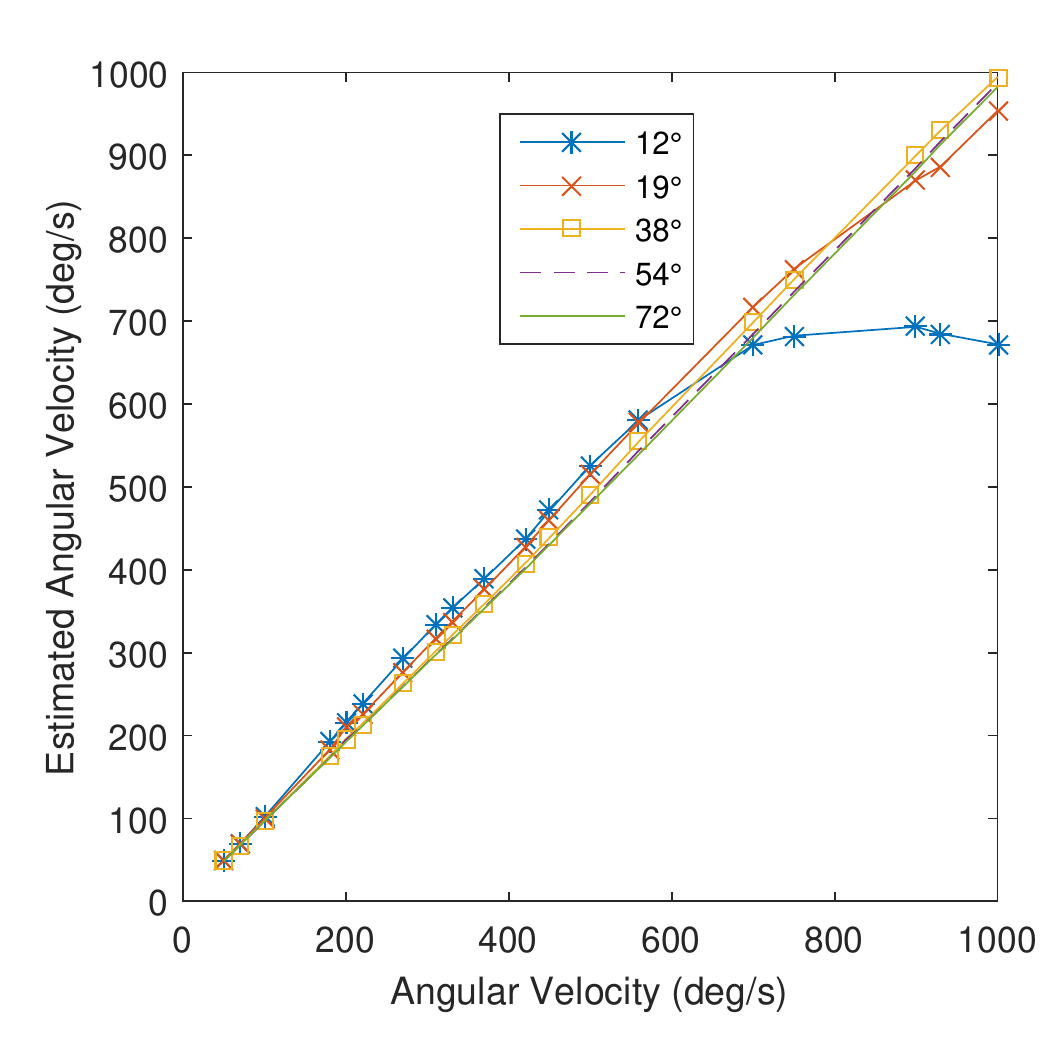}
\caption{ The estimated angular velocity curves from decoding under different angular velocities when tested by moving gratings of different spatial periods (12$^\circ$, 19$^\circ$, 38$^\circ$, 54$^\circ$ and 72$^\circ$).}
\label{fig4}
\end{figure}

As you can see from Fig. \ref{fig4}, the proposed model show indeed the expected independence of spatial frequency when tested by gratings of different spatial periods. The angular velocities are well decoded with little variance except when the grating is too narrow. This is caused by the much higher temporal frequency when the angular velocity is larger than 700$^\circ/s$ for grating of 12$^\circ$. Actually it does not affect the honeybee's flight in most of the cases since honeybees tend to maintain a constant angular velocity of 300$^\circ/s$ \cite{Baird2005}, around which our model shows pretty enough spatial independence.

\subsection{Terrain following simulations}
In the second kind of experiments, the proposed model is implemented in a simulated bee in terrain following simulations where the ground is covered with different textures. The flight trajectories and the ventral responses are recorded to see if the simulated bee can perform automatic terrain following by estimating the angular velocity of the image motion and regulating it to a constant value.

\begin{figure}
\centering
\includegraphics [width=120mm]{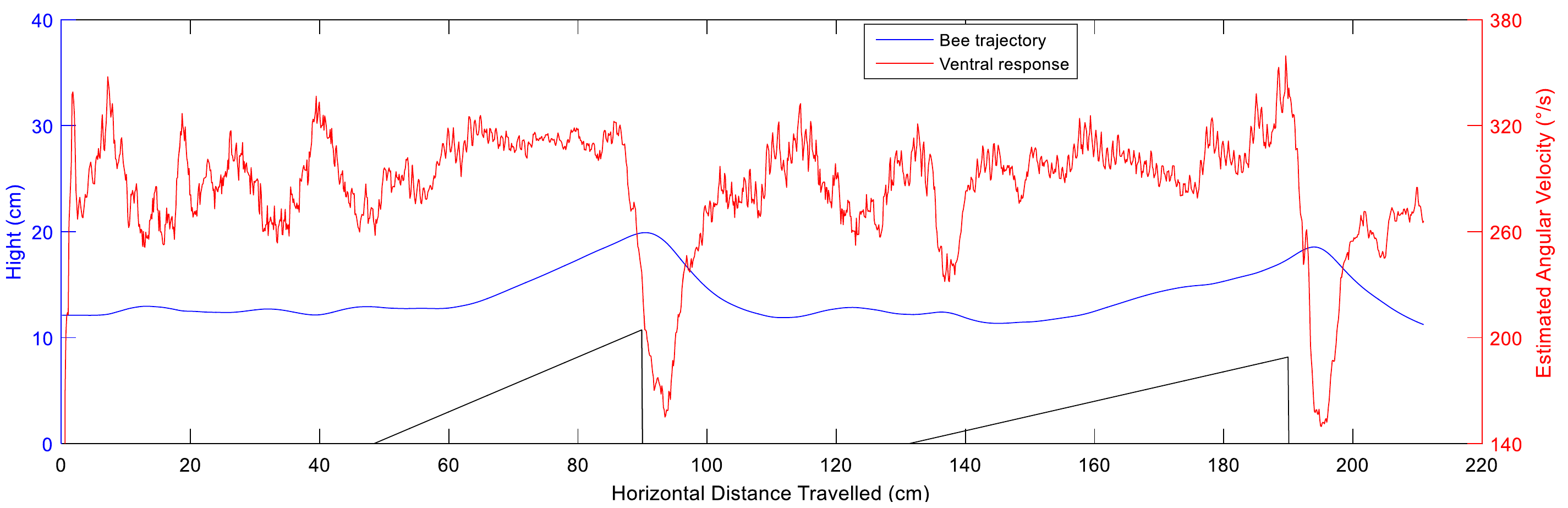}
\caption{ The bee trajectory (blue line), terrain height (black line) and the angular velocity (red line) estimated by the ventral eye are shown in the same graph. A demo video can be found at https://youtu.be/jaYSuCJGAfc.}
\label{fig5}
\end{figure}

The simulated bee with AVDM implemented is first tested on a regular terrain covered with sinusoidal gratings. The simulated bee is released around a given height at a certain forward speed. In beginning phase, the agent is set to fly forward without changing its altitude (by setting the vertical lift equals gravity), and the preset angular velocity value is estimated using AVDM after the first few frames. Then the control scheme described in previous section starts to take over the control of the vertical lift according to the difference between the angular velocity estimated and the preset value. The result is shown in Fig. \ref{fig5}.

From the flight trajectory, we can see the flight altitude changes automatically to keep a distance from ground using only visual information. This result is similar to the experiments on regular terrain with gratings \cite{ruffier2005optic, ruffier2015optic}. Using Unity engine, the images received by ventral camera can be processed in real time to estimate the image motion angular velocity. Neither the flight speed nor the flight altitude is necessary to perform this visual guided task.

\begin{figure}[hb]
\centering
\includegraphics [width=120mm]{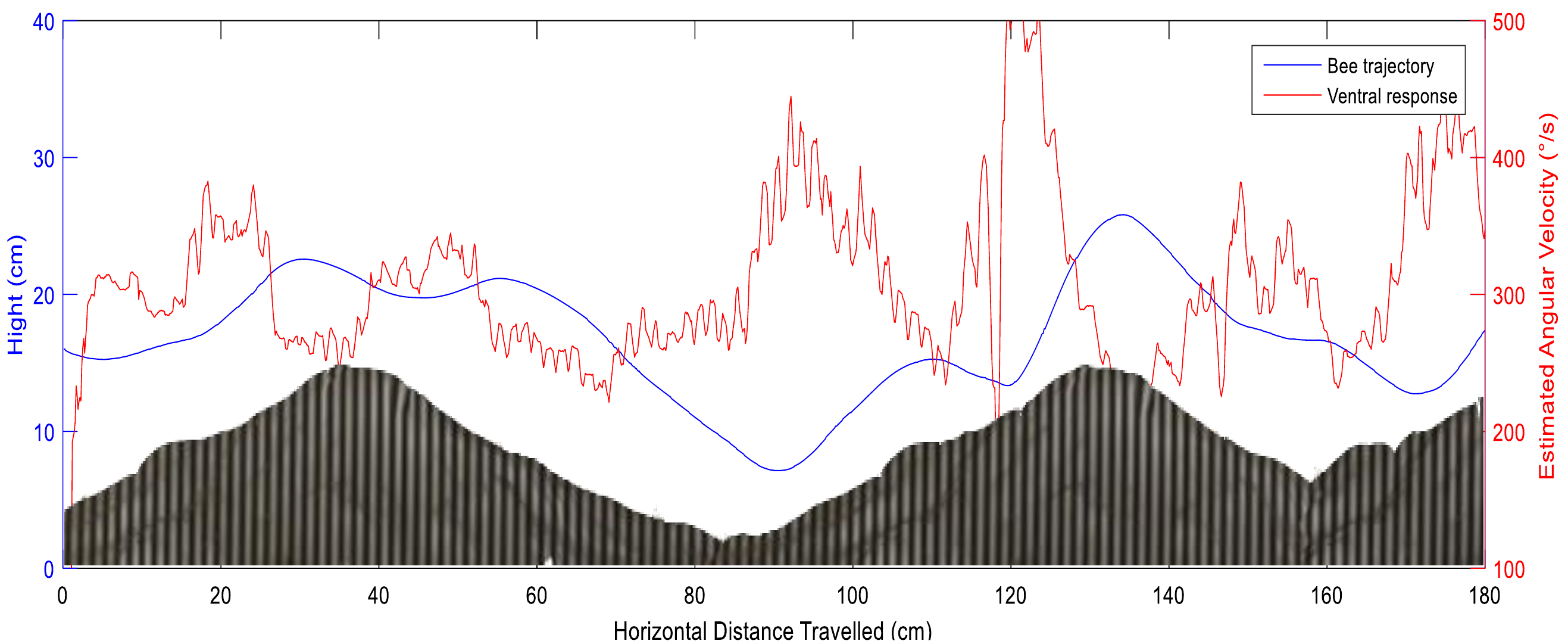}
\caption{ The bee trajectory (blue line) and the angular velocity (red line) estimated by the ventral eye are shown above the mountain terrain covered with gratings. The simulated bee can adjust its flight height by regulating the angular velocity to a constant value.}
\label{fig6}
\end{figure}

In order to see whether the model is stable under more complex terrain, we also tested the simulated bee above an irregular mountain terrain covered with sinusoidal gratings. The result is shown in Fig. \ref{fig6}. The flight trajectory of the agent undulates above the mountain shape ground which indicates that the flight altitude changes automatically according to the distance to ground using only visual information. Whenever the distance is close to ground, the increasing angular velocity will trigger the controller to provide a high vertical lift to help the agent get away from ground.

\begin{figure}
\centering
\includegraphics [width=120mm]{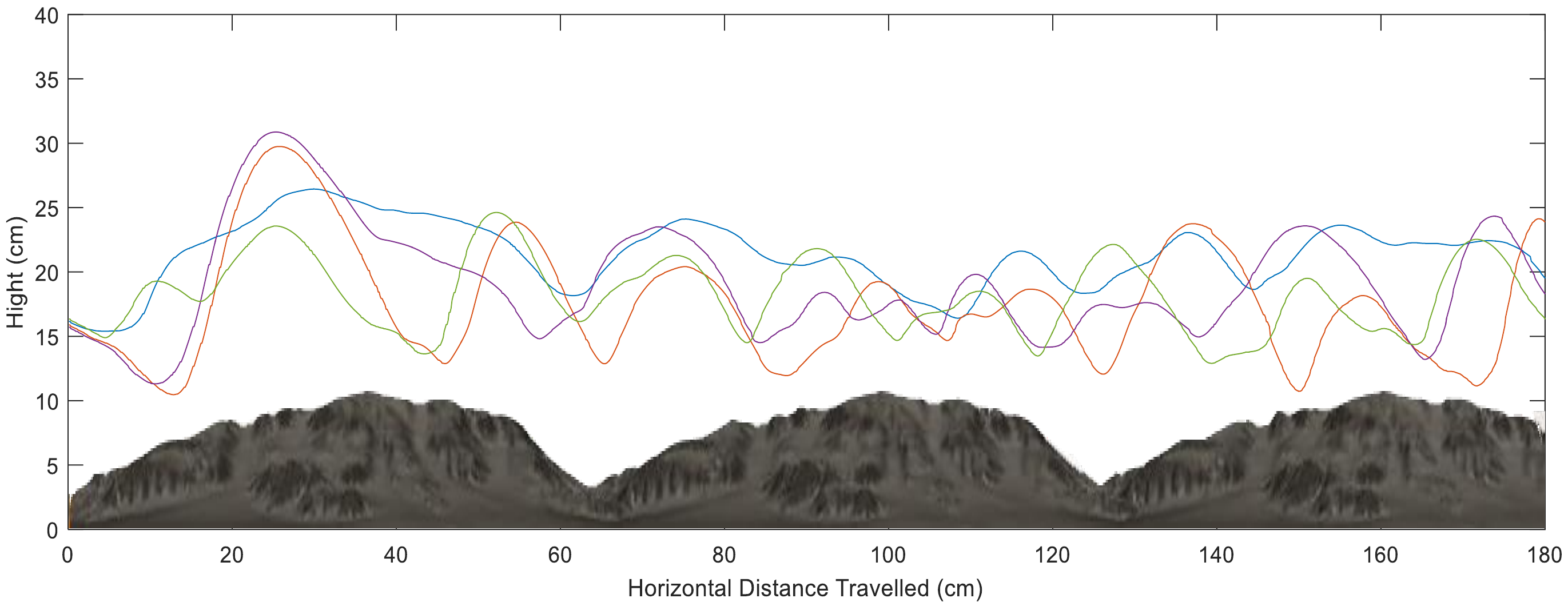}
\caption{ The simulated bee released around the given height in four times. The flight trajectories are shown above the snow mountain terrain in the same graph. Whenever the distance to ground decreases, the angular velocity increases causing the vertical lift goes up, and so does the flight altitude to help it fly over the terrain.}
\label{fig7}
\end{figure}

Further, the model is also tested using mountain terrain covered with white snow and black rock in our simulations. In this scenario, the texture information captured by ventral camera is irregular. Especially when there is too much snow in the vision field, the response will drop due to the low contrast. In order to inspect the terrain following ability in this situation, several simulations are performed (see Fig. \ref{fig7}). Though not all flight trajectories follow the terrain well, they all complete the flight tests without crashing. The result is obviously not robust as the terrain following using grating patterns. This is mainly caused by the lack of contrast and in accordance with the phenomenon that bees can plunge straight into calm water where low contrast provided \cite{heran1963}.

\section{Conclusion and Discussion}

We proposed a bio-plausible model, the angular velocity decoding model (AVDM), for estimating the image motion velocity to perform automatic terrain following with only visual information. The model combines both spatial and temporal information from moving frames received by ventral camera to give a relatively accurate image motion angular velocity. The response curves show large spatial independence in grating simulations which is in accordance with the biological experiments. It also provides a possible explanation of how insects' visual circuits detect image motion speed.

The close loop control scheme based on angular velocity estimation is also proposed to simulate the terrain following of honeybees in game engine Unity. And the result indicates that the model is well capable of guiding the flight course using only visual information. Similar strategy can be used in simulations of the honeybee's wall following behaviour \cite{serres2008bee}. The control scheme proposed mainly accounts for vertical lift regulating using only ventral part of the panoramic compound eye. In fact, similar strategy can be used to control forward thrust by lateral visual information. Combining both may give a more complete control scheme which can deal with more complex visual flight tasks. The proposed model will be implemented on UAV and be tested in a real environment in the near future.

\section{Acknowledgments}
 This research is funded by the EU HORIZON 2020 project, STEP2DYNA (grant agreement No. 691154) and ULTRACEPT (grant agreement No. 778062); the National Natural Science Foundation of China (grant agreement No. 11771347)

\bibliographystyle{splncs04}

\bibliography{Reference}

\end{document}